\documentclass[11pt]{article}

\usepackage[final]{acl}

\usepackage{times}
\usepackage{latexsym}

\usepackage[T1]{fontenc}

\usepackage[utf8]{inputenc}

\usepackage{microtype}

\usepackage{inconsolata}

\usepackage{graphicx}

\usepackage{multirow}

\usepackage{tikz}
\usetikzlibrary{positioning, shapes.callouts, fit}
\tikzset{every picture/.style={line width=0.75pt}} 
\usepackage{amsmath}
\usepackage{svg}
\usepackage{booktabs}
\usepackage{soul}
\usepackage{array}

\usepackage{todonotes}

\newcommand{\unrelated}{\emph{Unrelated}}

\newcommand{\Fmictarg}{$F_{\textrm{mic\_targ}}$}

\title{From Two Passes to One: Compact and Efficient Target-Stance Extraction}

\author{Ethan Mines, Bonnie Dorr \\
        University of Florida, Gainesville, FL, USA \\
        \texttt{\{ethanlmines, bonniejdorr\}@ufl.edu} }

\begin{document}
\maketitle
\begin{abstract}
Target-Stance Extraction (TSE) is the task of predicting both the target (or topic) of an author's writing and the author's stance toward it. Existing approaches to TSE use a sequential pipeline of two separate neural models: one to identify the target and another to determine the stance. We present a one-pass, joint architecture that predicts both in a single forward pass, reducing trainable parameters by nearly 50\% with only a 4--7 F1 point tradeoff in performance. We further demonstrate that standard target-scrubbing practices artificially suppress target prediction accuracy. Retaining explicit target mentions, as in real-world deployments, improves F1 by at least 6 points across both target classification and target generation settings. These improvements allow for significantly easier integration of TSE in downstream applications such as public opinion tracking.
\end{abstract}

\section{Introduction}

Stance detection is the task of classifying a document's stance towards a specific target.
A target may be a corporate merger \cite{conforti-etal-2020-will}, a political candidate \cite{li-etal-2021-p}, or a COVID-19 policy \cite{glandt-etal-2021-stance}.
In any case, the target must already be known 
to classify the document's stance.
This requires either human annotation of targets or narrowly restricting the corpus to documents known to 
refer to a particular target, potentially by keyword searches on social media \cite{conforti-etal-2020-will}.
Unfortunately, such prior knowledge is not always available.

The more general problem of Target-Stance Extraction (TSE), defined by \citet{li-etal-2023-new}, requires the system to predict both the target of the document and the stance toward that target.
The target prediction component of TSE includes two settings: Target Classification (TC) and Target Generation (TG).
Under TC, a classifier head is added to a pretrained language model (PLM) and used to predict among a fixed pool of targets.
In TG, a sequence-to-sequence model predicts a sequence of tokens representing the document's target.
In both cases, stance is predicted using a separate neural model taking the target as input.

\begin{figure}
    \centering
    \includegraphics[width=0.9\columnwidth]{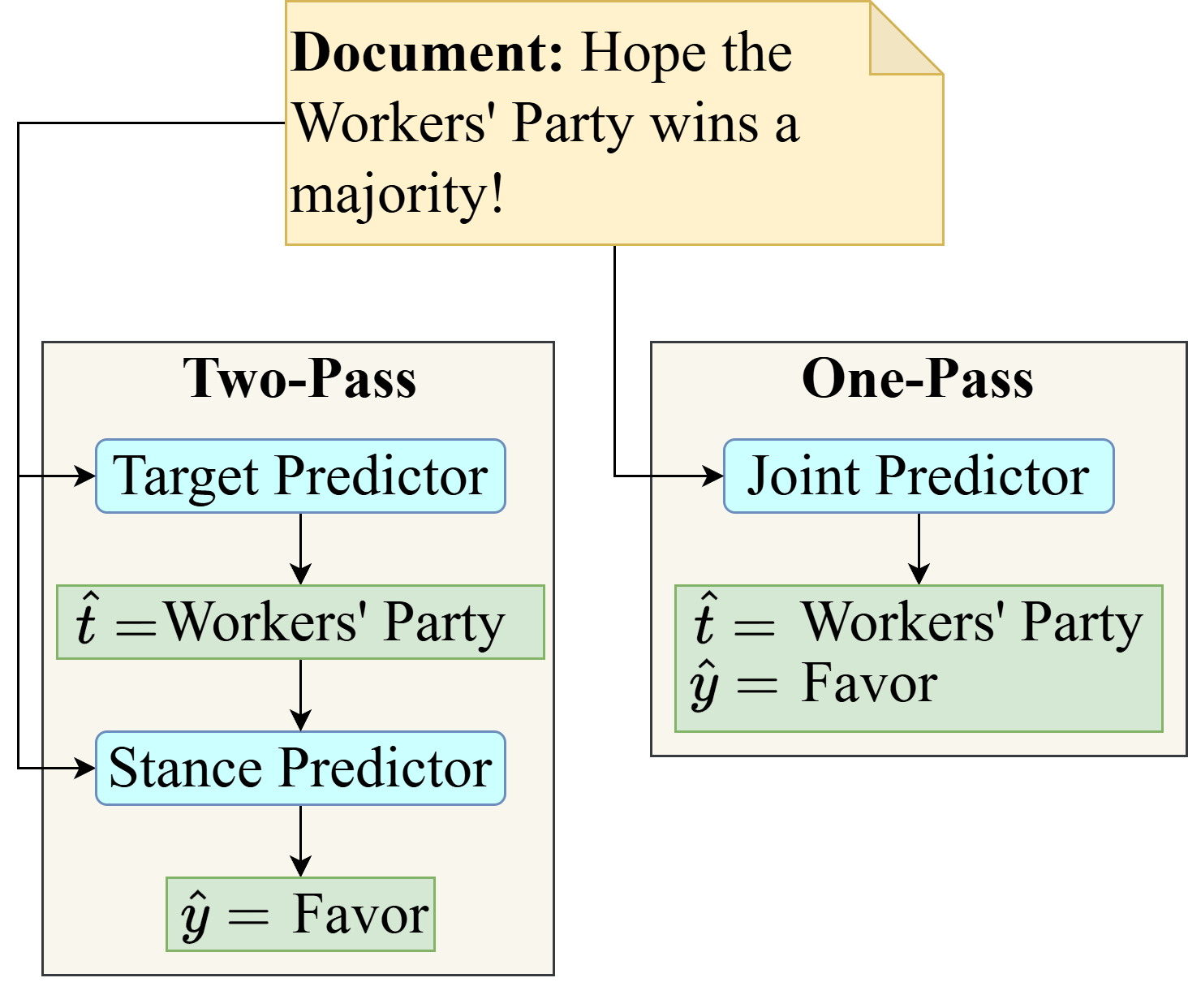}
    \vspace*{-.15in}
    \caption{Predicting a target $\hat{t}$ and a stance $\hat{y}$ with two passes (prior approaches) versus one (our approach).}
    \label{fig:two_vs_onepass}
    \vspace*{-.25in}
\end{figure}

In the object detection literature, R-CNN requires two passes over an image: one for region proposals and one 
for image classification \cite{rcnn}.
YOLO instead 
performs both in a single pass,
simplifying the pipeline and greatly improving efficiency \cite{yolo}.
As shown in Figure~\ref{fig:two_vs_onepass}, we 
follow this intuition
by condensing the two-model pipeline of \citet{li-etal-2023-new} into a single model 
that predicts target and stance, 
achieving comparable F1 scores while nearly halving the number of 
parameters.
This reduction 
not only improves inference efficiency but also simplifies deployment by consolidating all predictions 
within a single architecture.

Additionally, the original TSE benchmark provides the target predictor with data that have the target mentions removed.
While this may help the model learn implicit target cues, it does not reflect production environments, where explicit target cues are typically available and informative.
We show that retaining these mentions in the text can noticeably improve performance.  

Thus, the main contributions of our work are: (1) \textbf{reducing model footprint} while 
maintaining competitive performance, (2) \textbf{simplifying deployment} by replacing two models with one, and (3) \textbf{improving performance} by demonstrating the 
value of retaining explicit target cues.
Together, these enable more practical 
deployment for applications such as social media monitoring, public opinion tracking, and automated content analysis.\footnote{Our source code is available at: \url{https://github.com/elmines/multiling_tse}.}

\section{Related Work}

\S\ref{ssec:stance_det} covers prior research in stance detection, while \S\ref{ssec:tse} provides background on the emerging Target-Stance Extraction task.

\subsection{Stance Detection} \label{ssec:stance_det}

Stance detection is connected to the broader fields of Opinion and Argument Mining but is concerned with classifying the stance of an utterance towards a specific target (a consumer good, a political candidate, etc.).
Label inventories vary among datasets, but \{Favor, Against, None\} is common \cite{mohammad-etal-2016-semeval,allaway-mckeown-2020-zero,zhao-caragea-2024-ez}.

Early 
stance detection work focuses on online debate forums \cite{somasundaran-wiebe-2010-recognizing}.
As social media use has become more prevalent, the stance community has turned to classifying posts from sites such as Twitter \cite{mohammad-etal-2016-semeval} and SinaWeibo \cite{xu_overview_2016}.
Whereas these initial datasets limit possible target annotations to a small, fixed pool,
later benchmarks such as VAST \cite{allaway-mckeown-2020-zero} and EZ-Stance \cite{zhao-caragea-2024-ez} require zero-shot stance detection (ZSSD): predicting the stance towards an arbitrary target not necessarily seen in training or sample data.
 
Modern stance detection algorithms are typically machine learning models.
If the pool of possible targets is small, one can train a separate model for each target \cite{zarrella-marsh-2016-mitre}.
Alternatively, a model can accept a text representation of the target as part of its input \cite{augenstein-etal-2016-stance, xu-etal-2018-cross}.
It is common to fine-tune a pretrained BERT model \cite{devlin2019bert} for stance detection by providing it a concatenation of the target text and document text \cite{luo-etal-2022-exploiting, he-etal-2022-infusing, li-caragea-2021-multi}:

\begin{center}
\texttt{[CLS] target [SEP] document [SEP]}
\end{center}
A classifier head atop the model is then trained to predict the stance label.

As in 
previous work, we fine-tune pretrained language models like BERT and BART \cite{lewis-etal-2020-bart} for stance classification.
However, their methods require a target annotation as part of the input, whereas our algorithm requires only the document and predicts both target and stance.

\begin{figure*}[h]
    \centering
    \includegraphics[width=0.95\linewidth]{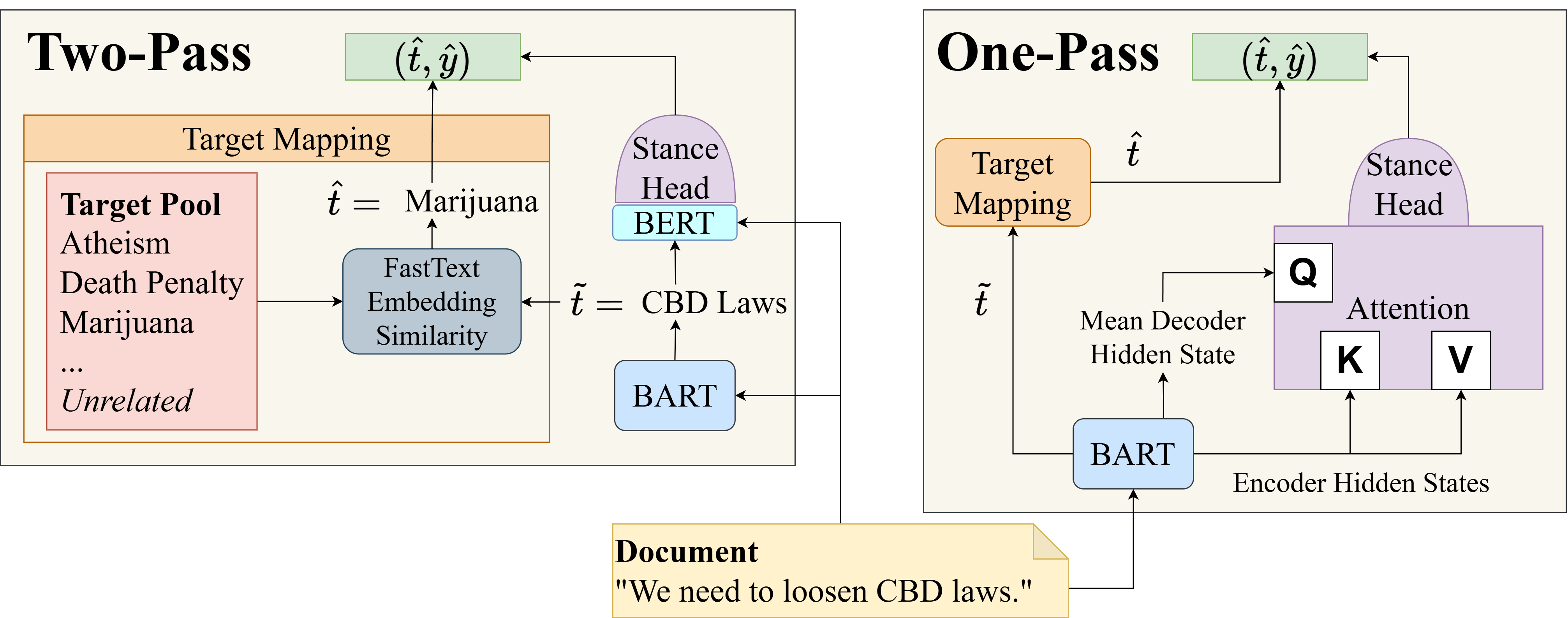}
    \caption{Comparison of Two-Pass and One-Pass TSE under the Target Generation (TG) scenario. In the two-pass setting,
    a BART sequence transduction model predicts the target while a separate BERT model predicts the stance. In the one-pass setting,
    a lightweight attention layer is appended to 
    BART for stance prediction. In both cases, FastText embeddings 
    map the free-form target prediction $\tilde{t}$ to a fixed target $\hat{t}$ for evaluation purposes.}
    \label{fig:target_gen}
\end{figure*}

\subsection{Target-Stance Extraction} \label{ssec:tse}

\citet{li-etal-2023-new} introduce the problem of Target-Stance Extraction: predicting both the target and stance given only the document.
They present two alternative scenarios: Target Classification (TC) and Target Generation (TG).
Under TC, a classifier head is added to a pretrained BERT model \cite{devlin2019bert}, which is then fine-tuned to predict among a fixed pool of targets.
For TG, a BART model \cite{lewis-etal-2020-bart} predicts a sequence of tokens, allowing for a free-form target.
They create a TSE benchmark using four existing stance detection corpora that are limited to eighteen targets \cite{mohammad-etal-2016-semeval,stab-etal-2018-cross,li-etal-2021-p,glandt-etal-2021-stance}.
In the case of TG, this requires them to map the predicted target tokens to a predetermined target from the evaluation set.

\citet{yan_bcltc_2025} improve on the TC scenario using prompt engineering and a special curriculum learning technique.
They still use a two-model pipeline, however.
\citet{akash-etal-2025-large} instead focus on the TG scenario and use a human evaluation of the predicted targets and stances, without mapping to predefined targets.
They prompt an LLM to predict both target and stance in a single pass.

Our work advances the TSE task by achieving competitive performance while reducing the number of models from two to one and, in turn,
cutting trainable parameters by nearly 50\%.
Our technique requires no prompt templates 
or curriculum learning.
The only existing work that performs one-pass TSE is that of \citet{akash-etal-2025-large}, who 
substantially \emph{increase} model footprint by relying on LLMs, whereas we \textit{reduce} model footprint while preserving competitive performance.
Moreover, their one-pass setting applies only to the TG scenario, while our approach demonstrates a unified one-pass architecture for both TC and TG.

Building on these prior methods in stance detection and TSE, the following section outlines our one-pass approach to TSE.

\section{Methodology}

The TSE problem is defined in \S\ref{ssec:prob_def}.
Our one-pass approaches to the TG and TC scenarios are described in \S\ref{ssec:tgen} and \S\ref{ssec:tcls}, respectively.

\subsection{Problem Definition} \label{ssec:prob_def}
Let $Y = \{\textrm{Against, Favor, None}\}$ denote
the stance label inventory and $T$ be the pool of possible targets.
A special target, $\textrm{\unrelated} \in T$, represents the absence of any meaningful target in the document.
Let $x$ be a document, $t \in T$ be a groundtruth (GT) target annotation, and $y \in Y$ be a GT stance annotation.
The goal of stance detection is to predict a stance $\hat{y} \in Y$ given $(x, t)$.
The goal of TSE is to predict a target-stance pair $(\hat{t}, \hat{y}) \in T \times Y$ given only $x$.

\subsection{Target Generation (TG)} \label{ssec:tgen}

As shown in Figure~\ref{fig:target_gen}, the two-pass approach of \citet{li-etal-2023-new} uses a BART model \cite{lewis-etal-2020-bart} for generating a target and a BERT model \cite{devlin2019bert} 
to classify stance.
BART is first pretrained on a keyphrase generation corpus such as KPTimes \cite{gallina-etal-2019-kptimes},
and its keyphrase predictions 
serve
as target predictions.
The BERT model then predicts stance using a concatenation of the BART model's token output $\tilde{t}$ and the original document.

To perform automatic evaluation of the free-form target $\tilde{t}$, it is mapped to a fixed target $\hat{t} \in T$ based on embedding similarity. 
Both $\tilde{t}$ and each target from the pool are mapped to a vector space using FastText embeddings \cite{bojanowski-etal-2017-enriching}.
If none of the cosine similarity values between $\tilde{t}$ and the pool's targets meets a threshold $\tau$, the \unrelated\ target is assigned, representing a negative prediction.
Otherwise, the prediction is mapped to the pool target with the highest cosine similarity.

Under our approach, a single BART model predicts both target and stance.
Target prediction and mapping follow the same procedure as in the two-pass setting.
For stance classification, we add a lightweight attention mechanism in which the mean of the BART decoder's hidden states serves 
as a query over the encoder's hidden states.
The output vector computed by the attention layer implicitly captures target cues, even  without direct access to the verbalized target tokens.
It serves as the feature vector given to the classifier head.
This novel architectural modification enables stance inference directly from decoder representations rather than from handcrafted target prompts.

Model training alternates between batches of keyphrase generation samples and batches of stance detection samples.
In the former case, language modeling loss for the keyphrases is minimized.
In the latter case, the GT target is provided to the BART decoder via teacher-forcing, and stance classification cross entropy is minimized.

Stance detection samples are never used to teach target prediction, as they would only bias the decoder toward generating targets in their fixed pool $T$.
The goal of TG on the other hand is prediction of diverse targets not necessarily in the pool.
Under TG, $T$ is only used in the target mapping stage during evaluation.

\subsection{Target Classification (TC)} \label{ssec:tcls}

\begin{figure}
    \centering
    \includegraphics[width=0.95\linewidth]{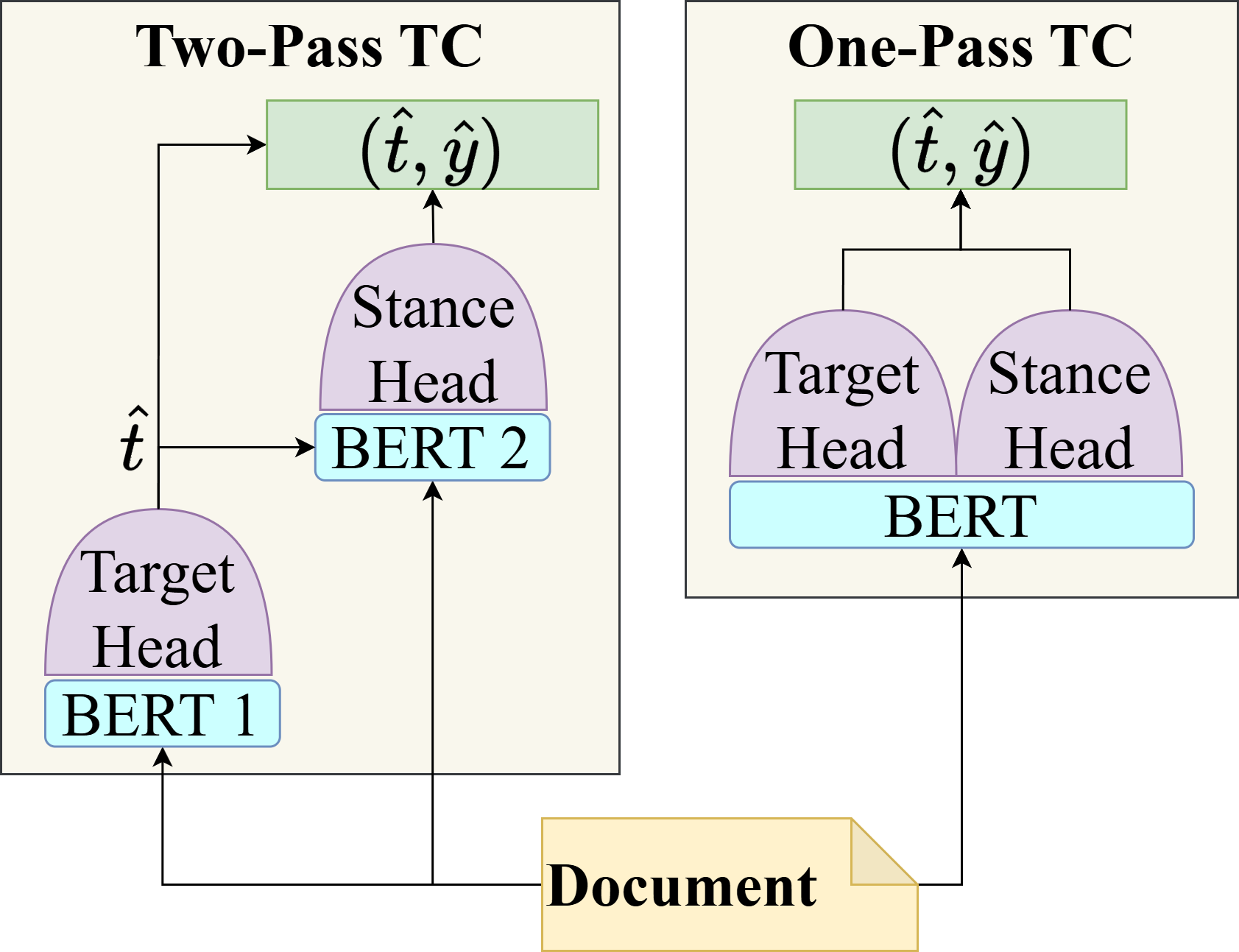}
    \caption{Comparison of Two-Pass and One-Pass TSE under the Target Classification (TC) scenario. In the two-pass variant, two separate models predict the target and stance. In the one-pass variant, one BERT model with two classifier heads predicts both.}
    \label{fig:target_cls}
\end{figure}

Under the existing approach of \citet{li-etal-2023-new}, two BERT models are used for TSE.
As shown in Figure \ref{fig:target_cls}, the first BERT model processes the document and predicts one target $\hat{t} \in T$.
The second processes a concatenation of the document and the target and predicts the stance.
Both models use the final hidden state of the \texttt{[CLS]} token as the feature vector for their classifier heads.
Each model is trained on cross entropy loss for its set of classes (targets or stances).
The stance model only sees GT targets during training.

Under our one-pass approach, a single BERT model is used for target and stance prediction.
Each classifier head receives the same \texttt{[CLS]} feature vector;
 thus, the evidence supporting target identification and stance is resolved jointly from the same document encoding, rather than sequentially conditioned on an intermediate target prediction.
This frames TSE as a multi-task learning problem within a single shared semantic space.

In both the two-pass and one-pass approaches, the models are trained on existing stance detection corpora with GT target and stance annotations.
In contrast to the TG scenario, here $T$ is needed for training in addition to evaluation, since the target head must predict from a fixed set of classes.

\subsection{Data}

\begin{table}
    \scriptsize
    \centering
    \begin{tabular}{p{0.10\linewidth}p{0.22\linewidth}p{0.50\linewidth}}
        \hline
        \textbf{Subset} & \textbf{\#Train}/\textbf{\#Val}/\textbf{\#Test} & \textbf{Unique Targets} \\
        \hline
         SE & 2160/359/1080 & Atheism, Feminist Movement, Hillary Clinton, Abortion \\
         \addlinespace[1.5pt] 
         AM & 18341/2042/5109 & Abortion, Cloning, Death Penalty, Gun Control, Marijuana Legalization, Minimum Wage, Nuclear Energy, School Uniforms \\
         \addlinespace[1.5pt]
         C19 & 4533/800/800 & Face Masks, Fauci, Stay at Home Orders, School Closures \\
         \addlinespace[1.5pt]
         PS & 17224/2193/2157 & Joe Biden, Bernie Sanders, Donald Trump \\
         \addlinespace[1.5pt] 
         Unrelated & 8800/1120/1880 & \unrelated \\
         \hline
    \end{tabular}
    \caption{Data partition sizes for SemEval-2016 Task 6 (SE), Argument Mining (AM), COVID-19 (C19), P-Stance (PS), and Unrelated components of the corpus.}
    \label{tab:dataset}
\end{table}

We use the same concatenated dataset aggregated by \citet{li-etal-2023-new} with its original train-validation-test split.
The source corpora include the SemEval-2016 Task 6 dataset \cite{mohammad-etal-2016-semeval}, an argument mining corpus \cite{stab-etal-2018-cross}, COVID-19-Stance \cite{glandt-etal-2021-stance}, and P-Stance \cite{li-etal-2021-p}.
Additional samples annotated with the \unrelated\ target are 
provided
by \citet{li-etal-2023-new}.
Dataset statistics 
are shown in Table~\ref{tab:dataset}.
51,000 samples from the KPTimes dataset \cite{gallina-etal-2019-kptimes} are used for the keyphrase training,
matching the approximately 51,000 samples used for stance training.
As in \citet{li-etal-2023-new}, the training fold of the stance corpus is used to learn the FastText embeddings.

All data are in English, drawn from social media or newswire sources, and purged of Personally Identifying Information (PII).
Since TSE involves the analysis of public discourse on often controversial issues, many of the stance samples by necessity contain offensive content.

\section{Experiments}

We evaluate both the existing two-pass algorithms of \citet{li-etal-2023-new} along with our one-pass versions.
Evaluation metrics are defined in \S\ref{ssec:eval}, different preprocessing approaches are described in \S\ref{ssec:preproc}, and model training procedures are outlined in \S\ref{ssec:models}.

\subsection{Evaluations} \label{ssec:eval}
Each system is evaluated for 
target prediction alone 
and for joint prediction of target and stance.

For target prediction, a micro-average of F1 scores across targets (\Fmictarg) is computed.
For overall TSE performance, the F1 metric defined by \citet{li-etal-2023-new} is adopted.
A sample is defined to be a negative if it is annotated with the \unrelated\ target and positive otherwise.
The system obtains a true positive for predicting the correct target and stance for a positive sample.
Predicting any target for a negative sample yields a false positive.
Predicting \unrelated\ for a positive sample yields a false negative.
Predicting the wrong target or stance for a positive sample yields both a false positive and a false negative.
From these definitions F1 score may be calculated.

We omit a direct analysis of stance classification performance because this would assume the system has an independent stance model that can accept a GT target as input---an assumption that holds only for two-pass systems. 
As
a proxy, we follow \citet{li-etal-2023-new} in calculating the F1 score 
in which the target prediction is replaced with the GT target.
This F1 score 
serves as a measure of the system's stance performance.

This replacement of predicted targets with GT ones is performed not only in the final output, but also in intermediate stages where appropriate.
In two-pass systems, the GT target text is fed directly into the stance classifier, and in the one-pass TG model, it is used to teacher-force the decoder.
The one-pass TC model, however, cannot use GT target text, since it never uses a textual representation of the target.
Importantly, this special evaluation is independent of preprocessing approaches such as Target Scrubbing (outlined next); it is applied uniformly under all preprocessing conditions.

\subsection{Preprocessing} \label{ssec:preproc}
For their TC systems, \citet{li-etal-2023-new} scrub all target mentions from the text during target prediction to force the model to learn implicit target cues.
For instance, since both ``Hillary Clinton'' and ``Bernie Sanders'' are in the target pool, they remove the strings ``Hillary,'' ``Clinton,'' ``Bernie,'' and ``Sanders'' from each post:

\emph{ \#\st{Hillary}\st{Clinton} just cannot accept the fact that \#America does not like her. \#\st{Hillary}onCNN \#\st{Bernie}\st{Sanders} \#SemST }

However, this approach does not reflect production environments, where explicit target mentions are naturally present and can be exploited.
In light of that, we introduce a variant \textbf{No Scrub} for our experiments that retains these mentions.

Additionally, nearly all the SemEval-2016 samples contain a hashtag ``\#SemST,'' a clear cue to only 
predict a target from SemEval's pool.
The source code of \citet{li-etal-2023-new} has a regular expression intended to remove this hashtag, but due to a casing error the regex has no effect.
We introduce a variant \textbf{No Hashtag} for our TC experiments which properly removes the hashtag.
This hashtag is not relevant to the TG scenario, where target prediction is learned from keyphrase data without the hashtag.

\subsection{Models} \label{ssec:models}

For the one-pass TC models, the 
learning rates for the two classifier heads and pretrained parameters are 1e-3 and 2e-5, respectively.
Training runs
for five epochs on 64-sample batches with a validation check each epoch. 
The model checkpoint with the best validation TSE F1 score is retained.
Training and evaluation together take 
roughly 30 minutes.

For the one-pass TG models, the learning rate for the language modeling head, stance classifier head, and new attention layer is set to 4e-5.
The remaining parameters have a learning rate of 1e-5.
Training alternates between KPTimes keyphrase data and stance corpora data, 
using 32-sample batches.
In the former, 
the model minimizes a language-modeling objective;
in the latter, it minimizes stance cross entropy.
The GT targets from the stance corpus are never used 
for language-modeling training; otherwise, the model will simply memorize 
the target set.
The validation TSE F1 score is 
computed every 500 batches.
Training 
stops after 
five checks with no improvement, and the model checkpoint with the highest score is retained.
Experiments take 
about one hour.

FastText embeddings of size 256 are trained for 500 epochs (about one hour) on the stance training samples using the Gensim library\footnote{\url{https://pypi.org/project/gensim/}}.
\citet{li-etal-2023-new} do not specify a cosine similarity threshold for mapping predicted targets to fixed ones,
but we found $\tau = 0.2$ sufficient
during preliminary tuning on validation data.

Training uses the AdamW optimizer 
\cite{adamw}.
BERT and BART 
are initialized 
from 
HuggingFace\footnote{\url{https://huggingface.co/}} models vinai/bertweet-base \cite{nguyen-etal-2020-bertweet} and facebook/bart-base \cite{lewis-etal-2020-bart}, respectively.

We 
reproduce
the two-pass experiments from \citet{li-etal-2023-new}.
For the TC setup,
we use their 
auxiliary target prediction objective for 
stance classifier training.
Their TG target prediction source code
is unavailable, but we otherwise use the same hyperparameters. 
The BART target predictor is trained 
on 1M
KPTimes samples and allowed to run for 24 hours to ensure sufficient learning.
A validation check is performed every 500 batches on KPTimes, 
retaining the checkpoint with the lowest language-modeling loss.

All hyperparameter selections are either adopted from \citet{li-etal-2023-new} or chosen based on TSE F1 score performance on the validation set.
All reported results are averaged across three random seeds.
Each experiment uses an Nvidia B200 GPU for BART training and inference and an Nvidia L4 GPU for all other neural computations.

\section{Results}

\begin{table}[h]
        \small
        \centering
        \begin{tabular}{lcc}
                \hline
                Algorithm & Two-Pass & One-Pass \\
                \hline
                TC \cite{li-etal-2023-new}         & 75.59          & - \\
                TC (Ours)                          & 76.19          & 76.23 \\
                \hspace{5pt}+No Hashtag            & 74.51          & 74.58 \\
                \hspace{5pt}+No Scrub              & \textbf{84.67} & \textbf{83.95} \\
                \hspace{5pt}+No Hashtag + No Scrub & 83.79          & 82.27 \\
                \hline
                TG \cite{li-etal-2023-new} & 48.31 & - \\
                TG (Ours) & 25.96 & 28.25 \\
                \hspace{5pt}+No Scrub & \underline{49.32} & \underline{45.36}
        \end{tabular}
        \caption{\Fmictarg\ performance by system for both Target Classification (TC) and Target Generation (TG) scenarios. Best result for TC is \textbf{bolded} while the best result for TG is \underline{underlined}.}
        \label{tab:targ_res}
\end{table}

Table~\ref{tab:targ_res} shows the results for target prediction performance.
For the TC scenario, our baseline 
demonstrably reproduces the performance of \citet{li-etal-2023-new}.
The ``No Scrub'' preprocessing condition performs best for both the Two- and One-Pass systems, reinforcing that explicit target mentions 
provide valuable and realistic signals.
Unsurprisingly, removing the SemEval hashtag cue hurts performance, regardless of target scrubbing.

Because our two-pass TG system with target scrubbing achieves substantially lower performance (25.96 F1) than that reported by \citet{li-etal-2023-new}, it is likely that target scrubbing is not applied in their TG setting, although their source code is not available to confirm this.
Our system 
without scrubbing slightly outperforms theirs (49.32 to 48.31), likely 
because of the 24-hour training time.
Across 
preprocessing conditions, the one-pass systems achieve comparable or 
slightly better performance than their two-pass counterparts.

Each system's performance on the overall TSE task when using predicted targets is shown in Table~\ref{tab:tse_res}.
The one-pass TC systems' F1 scores are consistently within 4 points of their two-pass counterparts.
Given \citet{li-etal-2023-new} did not scrub targets for TG, the gap between two- and one-pass systems for TG is at most 7 points (40.05 to 33.69).

In general, 
systems with the most favorable preprocessing (+No Scrub) achieve the best results for both target prediction and TSE.
Across the board, retaining target mentions boosts F1 by at least 6 points.
Since 
stance detection is 
largely solved 
on the TSE benchmark 
corpora, it is unsurprising that preprocessing which 
helps target prediction 
drives overall performance.

Following \citet{li-etal-2023-new}, we perform an evaluation where target predictions are replaced with GT targets.
Shown in Table~\ref{tab:gt_tse_res}, these metrics provide an upper bound by bypassing target prediction errors.
They additionally measure stance classification performance, the main limitation on performance under the GT setting.

The gap between two- and one-pass systems is wider here because the one-pass models cannot accept the GT target annotations as input the way the stance models of the two-pass systems can.
Regardless, the one-pass systems still achieve competitive scores.
Note target scrubbing has minimal impact here because only the target prediction stage benefits from explicit target mentions.

\begin{table}[h]
        \small
        \centering
        \begin{tabular}{lcc}
                \hline
                Algorithm                          & Two-Pass      & One-Pass        \\
                \hline
                TC \cite{li-etal-2023-new}         & 53.30          & -              \\
                TC (Ours)                          & 54.00          & 50.67          \\
                \hspace{5pt}+No Hashtag            & 52.94          & 49.39          \\
                \hspace{5pt}+No Scrub              & \textbf{60.89} & \textbf{59.41} \\
                \hspace{5pt}+No Hashtag + No Scrub & 60.73          & 58.41          \\
                \hline
                TG \cite{li-etal-2023-new} & 38.92 & -  \\
                TG (Ours) & 19.85 & 18.19 \\
                \hspace{5pt}+No Scrub & \underline{40.05} & \underline{33.69} \\
        \end{tabular}
        \caption{TSE F1 Scores given Predicted Targets. Best result for Target Classification (TC) is \textbf{bold}; best for Target Generation (TG) is \underline{underlined}.}
        \label{tab:tse_res}
\end{table}

\begin{table}[h]
        \small
        \centering
        \begin{tabular}{lcc}
                \hline
                Algorithm                          & Two-Pass      & One-Pass        \\
                \hline
                
                TC \cite{li-etal-2023-new}         & 75.28          &  -    \\
                TC (Ours)                          & \textbf{75.91} & 69.74 \\
                \hspace{5pt}+No Hashtag            & \textbf{75.91} & 69.48 \\
                \hspace{5pt}+No Scrub              & 75.27          & \textbf{72.96}           \\
                \hspace{5pt}+No Hashtag + No Scrub & 75.27          & 75.27  \\
                \hline
                TG \cite{li-etal-2023-new} & \underline{79.49} & - \\
                TG (Ours) & 75.91 & 64.13 \\
                \hspace{5pt}+No Scrub & 75.27 & \underline{70.51} \\
        \end{tabular}
        \caption{TSE F1 Scores given GT targets. Best result for Target Classification (TC) is \textbf{bold}; best for Target Generation (TG) is \underline{underlined}.}
        \label{tab:gt_tse_res}
\end{table}

\section{Analysis}

Beyond simplifying the training and deployment process, using a one-pass model greatly reduces the model footprint.
Table~\ref{tab:param_counts} 
reports the reduction in model parameters from using a one-pass system instead of a two-pass system.
In both the TC and TG scenarios, the number of parameters is 
reduced by nearly half. 

To illustrate the effect of target scrubbing on target prediction performance, samples with a high amount of scrubbing along with targets generated by the one-pass TG model are shown in Table~\ref{tab:scrubbing}.
The model can still predict relevant targets like ``income inequality'' but loses the ability to predict otherwise obvious named entities.

\begin{table}[t]
    \centering
    \begin{tabular}{lcc}
        \hline
        Scenario & Relative & Absolute \\
        \hline
        TC & 49.89\% & 135,505,171 \\
        TG & 48.32\% & 133,142,803
    \end{tabular}
    \caption{Reduction of model parameter count from two-pass to one-pass systems for the Target Classification (TC) and Target Generation (TG) scenarios.}
    \label{tab:param_counts}
\end{table}

Predictions for the one-pass TG model with no scrubbing are shown in Table~\ref{tab:sample_preds}.
For each sample the generated target $\tilde{t}$ is relevant to the original text, though errors occur in the mapping, e.g., the assignment of ``religion and churches'' to ``Feminist Movement''.
The P-Stance example indicates the desired case where TSE generates a target other than the explicit human annotation (``rourke \& amp kamala harris'') that is relevant to the original text while still producing a stance label (``Against'') appropriate for the generated target.

While most stance predictions match the label, this does not guarantee correctness with respect to the generated target shown to the end user.
For example, the stance prediction for the first document is ``Against,'' matching the GT, but the document is arguably in favor of ``religion and churches.''

\begin{table}[ht]
\scriptsize
\begin{tabular}{p{0.25in}>{\raggedright\arraybackslash}p{1.1in}p{0.4in}p{0.4in}}
        \hline
                                  &     & \multicolumn{2}{c}{Target} \\
                                          \cmidrule(lr){3-4}
        Corpus & Document & Generated & GT \\
        \hline
SE & @DesireeAaron @\st{Hillary}\st{Clinton} Sheets \st{Clinton} Ya got to love it \#\st{Hillary}onCNN \#\st{Hillary}\st{Clinton} \#SemST & donald trump & Hillary Clinton \\
AM & Proponents of a higher \st{minimum} \st{wage} state that the current federal \st{minimum} \st{wage} of \$ 7.25 per hour is too low for anyone to live on ; that a higher \st{minimum} \st{wage} will help create jobs and grow the economy ; that the declining value of the \st{minimum} \st{wage} is one of the primary causes of \st{wage} inequality between low - and middle-income workers ; and that a majority of Americans , including a slim majority of self-described conservatives , support increasing the \st{minimum} \st{wage} . & income inequality & minimum wage \\
C19 & @Morning\_\st{Joe} Well, you Morning \st{joe}, idiots President \st{Trump} was correct on using \#hydroxychloriquine for \#coronavirus and \#\st{Trump} did not need \#Dr\st{Fauci} to tell him to use it or not. \#mikabrzezinski is a Dope, who is 2nd fiddle to \st{Joe}. So get behind your Man? and \st{stay} there Mika. & bronzirus & fauci \\
PS & Former Vice President \st{Joe} \st{Biden} holds a community event in Muscatine, Iowa. \#\st{Joe}\st{Biden} \#\st{Biden} \#IowaCaucus @Team\st{Joe} @\st{Joe}\st{Biden} \#\st{Biden}s \#Iowa \#muscatine @jules1327 @JennyUwishuknew @jmill\_30 @morenoandreak @BlueTxBlue Live Link & primaries and caucuses & Joe Biden \\
Unrelated & Lovin' that jazz. https://t.co/AAf5cBucla & jazz & Unrelated \\
        \hline
\end{tabular}
\caption{Target-scrubbed samples with high numbers of explicit target mentions and the corresponding target predictions by the one-pass TG model.}
\label{tab:scrubbing}
\end{table}

\begin{table*}
\footnotesize

\begin{tabular}{p{0.35in}>{\raggedright}p{2.1in}>{\raggedright}p{0.75in}>{\raggedright}p{0.55in}p{0.50in}p{0.35in}p{0.35in}}
        \hline

         & & \multicolumn{3}{c}{Target} & \multicolumn{2}{c}{Stance} \\
         \cmidrule(lr){3-5} \cmidrule(lr){6-7}
        
                          {Corpus}    &
                          {Document}  &
                          {Generated} &
                          {Mapped}    &
                          {GT}        &
        \multicolumn{1}{c}{Predicted} &
        \multicolumn{1}{c}{GT}          \\
        \hline

        SE & My Jesus, I wish to do nothing but Thy most holy will. \#Saint Philip Neri \#Catholic \#SemST & religion and churches & Feminist Movement & Atheism & Against & Against \\
        
        \rule{0pt}{1.35em}AM &  Abortions cause psychological damage .  & mental health & abortion & abortion & Against & Against \\
        
        \rule{0pt}{1.35em}C19 & My wife and I were not allowed to work as they say we now have to wear a mask. When told we are exempt they said we need letter from doctor to prove it. Rang doctor's and they say they will not give us a letter or refer us privately for one. \#KBF & mask & face masks & face masks & Neutral & Against \\
        
        \rule{0pt}{1.35em}PS & How about @BetoORourke \&amp; @KamalaHarris ticket? Not sure if @JoeBiden has been around too long??? & rourke \& amp kamala harris & Joe Biden & Joe Biden & Against & Against \\
        
        \rule{0pt}{1.35em}Unrelated & @HellNaw4Life @iamDOAS Hispanics \&amp; Latinos are already white & hispanic americans & Donald Trump & Unrelated & Neutral & Neutral \\

        \hline
\end{tabular}
\caption{Randomly selected samples without target scrubbing and corresponding predictions by the one-pass Target Generation (TG) model.}
\label{tab:sample_preds}
\end{table*}

\section{Conclusion}

This work introduces the first one-pass architecture for both the TC and TG scenarios of Target-Stance Extraction. By unifying target and stance prediction in a single model, we reduce model size by nearly 50\% while 
maintaining competitive performance.

More importantly, we demonstrate that existing target-scrubbing practices systematically suppress performance.
Retaining explicit target mentions, as in naturally occurring texts, leads to consistent improvements of at least 6 F1 points, suggesting that more realistic preprocessing is critical for trustworthy evaluation.

Together, these findings shift the focus of TSE from complex multi-model pipelines toward simple and efficient architectures that are more easily deployed in practical applications.

\section{Future Work}

Avenues for future work include (a) \textit{predicting the target before stance in the one-pass model}, (b) \textit{using zero-shot stance detection (ZSSD) corpora for TSE}, and (c) \textit{predicting multiple target-stance pairs for a single document}.

\textbf{(a) Predicting the target before stance in the one-pass model.}
Both the one-pass TC and TG models pass the document through the backbone (BART or BERT) and make a final prediction for both target and stance following the backbone.
The target could be predicted using only the first few layers of the backbone, and this prediction could serve as additional input to subsequent layers.
Such an approach could improve the accuracy of the stance classification step by providing additional context from the target, while still limiting the TSE pipeline to a single neural model.
Additionally, this retains the interpretability of the two-pass approach by predicting the target and stance in sequence.

\textbf{(b) Using zero-shot stance detection (ZSSD) corpora for TSE.}
The stance corpora used by \citet{li-etal-2023-new} in total have less than 20 possible targets.
While annotating a new corpus specifically for TSE would be ideal, finding a way to incorporate existing ZSSD datasets containing thousands of different targets \cite{allaway-mckeown-2020-zero, zhao-etal-2023-c, zhao-caragea-2024-ez} would allow for better training and evaluation of TSE systems.

\textbf{(c) Predicting multiple target-stance pairs for a single document.} Stance detection is generally limited to predicting the stance for a single target; however, a single document may mention 
multiple targets and 
express different stances toward each one.
A TSE algorithm that could detect all targets and their corresponding stances would both give users a better summary of an individual document while also allowing them to analyze the co-occurrence of target-stance pairs across an entire corpus.

\section*{Limitations}

While we follow \citet{li-etal-2023-new} in the use of cosine similarity with FastText embeddings for mapping generated targets to the pool's fixed labels, this heuristic only approximates human judgment and can introduce evaluation mismatches.
Both the TC and TG scenarios require a fixed target pool for evaluation.
Future work could explore larger, dynamic, or even unbounded target pools, and evaluate how well one-pass architectures generalize to them.
Experiments described herein are limited to English data with a predominantly political focus; future work can explore other languages and domains.

\section*{Ethical Considerations}
All datasets used in the study are publicly available; to abide by social media sites' user privacy policies, we do not redistribute these data in any fashion.

TSE helps good actors derive data-driven insights about public opinion, but such insights could also enable bad actors to identify dissenting voices and target them for censorship or harassment.
As TSE algorithms advance in capability, restrictions on government and commercial use may become necessary to curb such misuse.
As a technological safeguard, adversarial algorithms \cite{zhao-caragea-2026-stanceattack} could help protect authors by modifying document content to fool TSE systems without altering the text's meaning.

\bibliography{anthology,TSEPaper}

\end{document}